\documentclass[letterpaper, 10 pt, conference]{ieeeconf}  

\IEEEoverridecommandlockouts                              

\usepackage{graphics} 
\usepackage{epsfig} 
\usepackage{mathptmx} 
\usepackage{times} 
\usepackage{amsmath} 
\usepackage{amssymb}  
\usepackage{bm}
\usepackage{algorithm}
\usepackage{algorithmic}
\usepackage{lipsum}

\DeclareMathOperator*{\argmin}{argmin}

\title{\LARGE \bf
Accelerating Model-Based Reinforcement Learning using Non-Linear Trajectory Optimization
}

\author{Marco Cali'$^{1}$, Giulio Giacomuzzo$^{1,2}$, Ruggero Carli$^{1}$, Alberto Dalla Libera$^{1,2}$
\thanks{$^{1}$Marco Cali', Giulio Giacomuzzo, Ruggero Carli and Alberto Dalla Libera are with Department of Information Engineering of University of Padua, Via Gradenigo 6, Padua, Italy 
{\tt\small marco.cali.2@studenti.unipd.it}}%
\thanks{$^{2}$Alberto Dalla Libera and Giulio Giacomuzzo were supported by PNRR research activities of the consortium iNEST (Interconnected North-East Innovation Ecosystem) funded by the European Union Next GenerationEU (Piano Nazionale di Ripresa e Resilienza (PNRR) – Missione 4 Componente 2, Investimento 1.5 – D.D. 1058  23/06/2022, ECS\_00000043). This manuscript reflects only the Authors’ views and opinions, neither the European Union nor the European Commission can be considered responsible for them.}%
}

\begin{document}

\maketitle
\thispagestyle{empty}
\pagestyle{empty}


%

\begin{abstract}


This paper addresses the slow policy optimization convergence of Monte Carlo Probabilistic Inference for Learning Control (MC-PILCO), a state-of-the-art model-based reinforcement learning (MBRL) algorithm, by integrating it with iterative Linear Quadratic Regulator (iLQR), a fast trajectory optimization method suitable for nonlinear systems. The proposed method, Exploration-Boosted MC-PILCO (EB-MC-PILCO), leverages iLQR to generate informative, exploratory trajectories and initialize the policy, significantly reducing the number of required optimization steps. Experiments on the cart-pole task demonstrate that EB-MC-PILCO accelerates convergence compared to standard MC-PILCO, achieving up to $\bm{45.9\%}$ reduction in execution time when both methods solve the task in four trials. EB-MC-PILCO also maintains a $\bm{100\%}$ success rate across trials while solving the task faster, even in cases where MC-PILCO converges in fewer iterations.

\end{abstract}
\section{INTRODUCTION}
In recent years, Reinforcement Learning (RL) has emerged as a promising technology for deriving controllers for complex tasks across different domains, ranging from virtual environments to Robotics. Unlike classical control theory, which typically relies on models derived from first principles, RL derives the controller, also named policy, in a data-driven fashion. 
The perspective of a system that learns autonomously from experience makes RL an attractive alternative to traditional control methodologies. However, RL is not without its challenges. 
Key among these, in the Robotics domain, is the sample efficiency, namely, the amount of data the algorithm needs to converge to a good solution. Indeed, in actual systems data collection is a time-consuming and safety-critical operation. 

MBRL \cite{polydoros2017survey} has been proposed as a data-efficient alternative to traditional Model-Free RL (MFRL) \cite{alphaGo}. Unlike MFRL algorithms, which estimate the value of an input directly from data, MBRL increases the information extracted from interactions by constructing a predictive model of the system. The policy is optimized on the model instead of the system, thus limiting experience on the actual system and potentially dangerous interactions. An interesting class of MBRL algorithms relies on Gaussian Process Regression (GPR) \cite{williams2006gaussian} to derive the system dynamics \cite{deisenroth2011pilco, MCPILCO, gal2016improving, cutler2015efficient, chatzilygeroudis2017black,parmas2018pipps}. Among these solutions, Monte Carlo Probabilistic Inference for Learning COntrol (MC-PILCO)  \cite{MCPILCO} is a state-of-the-art algorithm that proved particularly data-efficient in several actual applications \cite{competition, VFMCPILCO}. 
GPR, thanks to its Bayesian formulation, returns a stochastic model that accounts for intrinsic and epistemic uncertainty, namely due to limited data availability. This fact opens the possibility of considering uncertainty during policy optimization and deriving more robust controllers. These advantages come at the price of computationally intensive and time-consuming policy optimization stages, that could limit the applicability of such algorithms from a technological point of view. 
In this work, we investigate the combination of MC-PILCO with iLQR\cite{ilqr} to reduce MC-PILCO's convergence time. iLQR is a differential dynamic programming method that extends linear quadratic controllers to deterministic nonlinear systems by linearizing the dynamics around the current trajectory. Contrary to MC-PILCO, iLQR converges to a solution much more quickly. However, this solution is potentially local and less robust since it does not consider stochasticity. The main idea is that MC-PILCO convergence time can be reduced by using iLQR to explore the system in the neighborhood of low-cost regions and initialize the MC-PILCO policy, thus providing a warm start for MC-PILCO. To implement this strategy, we focused on two main aspects: (i) integration of Gaussian Process (GP) models in the iLQR framework \cite{agpilqr}, (ii) definition of an effective strategy to pretrain the MC-PILCO policy from an iLQR controller. We evaluated the proposed approach on the cart-pole swing-up task through extensive simulations. The results demonstrate faster convergence compared to MC-PILCO, highlighting its potential for broader applications. We named the resulting framework Exploration-Boosted MC-PILCO (EB-MC-PILCO).

\section{BACKGROUND}
This section reviews background notions on iLQR, GPR for dynamics model identification and the MC-PILCO algorithm. In the following, we consider a discrete-time dynamics system and point out with $\bm x_t \in \mathbb{R}^{d_x}$ and  $\bm u_t  \in \mathbb{R}^{d_u}$ its state and input at time t. We assume the following dynamics equation,
\begin{equation}
    \bm x_{t+1} = \bm f(\bm x_t, \bm u_t). \label{eq:transition}
\end{equation}
Without loss of generality, we consider a mechanical system, and, as usual, we assume that the system state is given by $\bm x_t = [\bm q_t^\top, \bm {\dot q}_t^\top]^\top$, where $\bm q_t$ and $\dot{\bm{q}}_t$ are the position and velocity of the system.

\subsection{iLQR}\label{sec:iLQR}
Iterative Linear Quadratic Regulator (iLQR) \cite{ilqr} optimizes control sequences for nonlinear systems by iteratively linearizing dynamics and solving a quadratic cost minimization problem. The goal is to minimize the cumulative cost:
\begin{equation}
    J(\bm x_0, U) = \sum_{t=0}^{T-1} c(\bm x_t, \bm u_t) + c_f(\bm x_T),
\end{equation}
where the running cost $c(\bm x_t, \bm u_t)$ and terminal cost $c_f(\bm x_T)$ are quadratic:
\begin{subequations}
\begin{align}
    c(\bm x_t, \bm u_t ) & = \frac{1}{2} \bm x_t^\top  Q \, \bm x_t +\frac{1}{2}  \bm u_t^\top  R \,\bm u_t, \\
    c_f(\bm x_T) & = \frac{1}{2} \bm x_T^\top  Q_f \, \bm x_T.
\end{align}
\end{subequations}
Here $ Q \in \mathbb{R}^{d_x \times d_x}$ and $Q_f\in \mathbb{R}^{d_x \times d_x}$ are the state cost matrices and $ R \in \mathbb{R}^{d_u \times d_u}$ is the input cost matrix. The algorithm starts from a nominal trajectory $\bm \zeta$, generated by applying the control sequence $U$ to the initial state $\bm x_0$. If a good initialization is not known, then we can set $U = 0$ \cite{Li_Todorov}. The algorithm consists of two main steps, detailed below.

\subsubsection{Backward Pass}
The backward pass computes the optimal control modification using the $Q$-function, which approximates the cost-to-go $J_\tau$:
\begin{align}
    J_\tau(\bm x, U_\tau)  & = \sum_{t=\tau}^{T-1} c(\bm x_t, \bm u_t) + c_f(\bm x_T) , \\
    Q(\delta \bm x, \delta \bm u) &=  c(\bm x + \delta \bm x , \bm u + \delta \bm u) + V'(\bm f(\bm x + \delta \bm x, \bm u +\delta \bm u))
\end{align}
where $V(\bm x) = \min_{\bm u} \left[ c(\bm x, \bm u) + V'(\bm f(\bm x, \bm u))\right]$ and $V'$ is the value function at the next time step, and $\delta \bm x = \hat {\bm x}- {\bm x}$ and  $\delta \bm u = \hat {\bm u}- {\bm u}$ are the deviations from the nominal trajectory. The $Q$-function is expanded quadratically around the nominal trajectory, and its derivatives are computed:
\begin{subequations}
\begin{align}
    Q_{\bm{x}} &= c_{\bm{x}} + f_{\bm{x}}^\top V_{\bm{x}}' \label{eq:Q_x}, \\
    Q_{\bm{u}} &= c_{\bm{u}} + f_{\bm{u}}^\top V_{\bm{x}}' \label{eq:Q_u}, \\
    Q_{\bm{xx}} &= c_{\bm{xx}} + f_{\bm{x}}^\top V_{\bm{xx}}' f_{\bm{x}} \label{eq:Q_xx} ,\\
    Q_{\bm{uu}} &= c_{\bm{uu}} + f_{\bm{u}}^\top V_{\bm{xx}}' f_{\bm{u}} \label{eq:Q_uu}, \\ 
    Q_{\bm{ux}} &= c_{\bm{ux}} + f_{\bm{u}}^\top V_{\bm{xx}}' f_{\bm{x}} \label{eq:Q_ux}. 
\end{align}
\end{subequations}
The optimal control modification is given by
\begin{equation}
    \delta \bm u ^\star = \argmin_{\delta \bm u} Q(\delta \bm x, \delta \bm u) = \bm k +  K \delta \bm x, 
\end{equation}
with open-loop term $\bm k \triangleq - Q_{\bm {uu} }^{-1} Q_{\bm u }$ and feedback gain term $K \triangleq - Q_{\bm {uu} }^{-1} Q_{\bm {ux} }$. Finally, the value function and its derivatives are updated for the next time step:
\begin{subequations}
\begin{align}
    V_{\bm x} & = Q_{\bm x} - K^\top Q_{\bm {uu}} \bm k  = Q_{\bm x} - Q_{\bm{ux}}^\top Q_{\bm {uu} }^{-1}Q_{\bm {u} }, \\
    V_{\bm {xx}} & = Q_{\bm {xx}} - K^\top Q_{\bm {uu}}  K = Q_{\bm {xx}} - Q_{\bm{ux}}^\top Q_{\bm {uu} }^{-1}Q_{\bm {ux} }
\end{align}
\end{subequations}
To ensure stability, $Q_{\bm{uu}}^{-1}$ is regularized via eigendecomposition. Small eigenvalues are increased by a regularization parameter $\gamma$, which is lowered as cost improves.

\subsubsection{Forward rollout}
The new trajectory $\bm \zeta$ is computed in the forward pass: 
\begin{align}
    \hat {\bm x}_0 & = \bm x_0,\\
    g(\hat {\bm x}_t) = \hat {\bm u}_t & = \bm u_t + \alpha \bm k_t + K_t (\hat {\bm x}_t - \bm x_t), \\
    \hat {\bm x}_{t+1} & = \bm f( \hat {\bm x}_t, \hat {\bm u}_t).
\end{align}
The backtracking parameter $\alpha$, initially set to 1, ensures cost improvement by selecting the largest possible step size. Hyperparameters $\alpha_f$ (step size reduction rate) and $\alpha_\text{min}$ (minimum step size) control convergence: higher  $\alpha_f$ speeds up convergence but risks oscillation, while lower $\alpha_\text{min}$ allows broader exploration but may slow convergence.

\subsection{GPR for dynamics model identification}\label{sec:gpr}
In the MBRL algorithm considered in this work the transition function \eqref{eq:transition} is identified from data by means of GPR. Function $\bm f(\bm x, \bm u)$ is reconstructed through $d_x$ independent GPs with input  $\tilde{\bm{x}}_t = [\bm{x}_t^\top \; \bm{u}_t^\top]^\top$, hereafter pointed out by $\Delta^{(i)}(\tilde{\bm{x}}_t)$ with $i=1\dots d_x$. The $i$-th GP $\Delta^{(i)}$ models the delta between the $i$-th components of consecutive states, namely
\begin{equation}
    \Delta^{(i)}(\tilde{\bm{x}}_t) = \bm{x}_{t+1}^{(i)} - \bm{x}_t^{(i)}.\label{eq:output-def}
\end{equation}
By definition of GP, $\Delta^{(i)}(\cdot) \sim GP(m(\cdot), k(\cdot,\cdot))$
where $m(\cdot)$ is the prior mean, while the kernel function $k(\cdot,\cdot)$ is a function that defines the prior covariance between $\Delta^{(i)}(\tilde{\bm{x}}_t)$ and $\Delta^{(i)}(\tilde{\bm{x}}_p)$, that is, $k(\tilde{\bm{x}}_t,\tilde{\bm{x}}_p) = Cov[\Delta^{(i)}(\tilde{\bm{x}}_t), \Delta^{(i)}(\tilde{\bm{x}}_p)]$. In the following, we will adopt the compact notation $ K(\tilde{{X}}, \tilde{{X}})$ to point out the kernel matrix associated to a set of input locations $\tilde{{X}}$. Since we assume no prior knowledge about the system, we set the GP' prior mean $m_i(\bm x) = 0$ and we adopt a Squared Exponential (SE) kernel expressed as
\begin{equation}
    k_i(\tilde {\bm x}_t, \tilde {\bm x}_p) = k_\text{SE}(\tilde {\bm x}_t, \tilde {\bm x}_p) := \lambda ^ 2 \exp\left(- ||\tilde {\bm x}_t - \tilde {\bm x}_p||^2_{\Lambda^{-1}}\right),
    \label{eq:SE_kernel}
\end{equation}
where $\lambda$ is a scaling factor and $\Lambda$ is the lengthscales matrix. For convenience, we define $\Lambda = \text{diag}(\Lambda_{\bm{x}}, \Lambda_{\bm{u}})$, where $\Lambda_{\bm{x}} \in \mathbb{R}^{d_x \times d_x}$ and $\Lambda_{\bm{u}} \in \mathbb{R}^{d_u \times d_u}$ are the diagonal matrices corresponding to $\bm{x}$ and $\bm{u}$, respectively. Let $D^{(i)}=\{\tilde{{X}}, \bm{y}^{(i)}\}$ be an input-output dataset, where $\bm{y}^{(i)}$ is a vector that collects $N$ noisy measurements of \eqref{eq:output-def} at the input locations $\tilde{{X}}=\{\tilde {\bm x}_1, \dots \tilde {\bm x}_N\}$; we assume that measures are corrupted by a zero mean additive Gaussian noise with standard deviation $ \sigma_i$. In GPR, $p(\Delta^{(i)}(\tilde {\bm x}_*)  | D^{(i)})$, the posterior distribution $\Delta^{(i)}$ in a general input location $\tilde {\bm x}_*$, is Gaussian with mean and variance given in closed form as
\begin{subequations}
\begin{align}
    \mathbb{E}\left[\hat{\Delta}_*^{(i)}\right] &= k_i(\tilde{\bm{x}}_*, \tilde{{X}}) \Gamma_i^{-1} \bm{y}^{(i)}, \label{eq:delta-posterior-mean} \\
    \text{Var}\left[\hat{\Delta}_*^{(i)}\right] &= k_i(\tilde{\bm{x}}_*, \tilde{\bm{x}}_*) - k_i(\tilde{\bm{x}}_*, \tilde{{X}}) \Gamma_i^{-1} k_i^\top(\tilde{\bm{x}}_*, \tilde{{X}}),\label{eq:delta-posterior-var}
\end{align}    
\end{subequations}
where $\Gamma_i = K_i(\tilde{{X}}, \tilde{{X}}) + \sigma_i^2 I$, and $k_i(\tilde{\bm{x}}_t, \tilde{{X}})$ is the row vector concatenating the kernel evaluations between $\tilde{\bm{x}}_*$ and training data \cite{williams2006gaussian}. Then, due to the properties of Gaussian distribution, also the posterior distribution of the next state given $\tilde {\bm x}_*$ and $ D^{(i)}$ is Gaussian, namely, $\hat{\bm{x}}_{t+1} \sim \mathcal{N}(\bm{\mu}_{t+1}, \Sigma_{t+1})$, where
\begin{subequations}
\begin{align}
    \bm{\mu}_{t+1} &= \bm{x}_t + \begin{bmatrix}
        \mathbb{E}[\hat{\Delta}_t^{(1)}] & \dots & \mathbb{E}[\hat{\Delta}_t^{(d_x)}]
    \end{bmatrix}^\top, \label{eq:one-step-ahead-posterior-mean}\\
    \Sigma_{t+1} &= \text{diag}\left(\begin{bmatrix}
        \text{Var}[\hat{\Delta}_t^{(1)}] & \dots & \text{Var}[\hat{\Delta}_t^{(d_x)}]
    \end{bmatrix}\right). \label{eq:one-step-ahead-posterior-var}
\end{align}
\end{subequations}
    
In the case that the state is $\bm x_t = [\bm q_t^\top, \bm {\dot q}_t^\top]^\top$, the \textit{speed-integration model} is an interesting option, proposed in \cite{MCPILCO} to reduce the computational burden and increase physical consistency. This model defines a GP for each speed component and obtains positions by integration, thus halving the number of GPs. Indeed, if the sampling time $T_s$ is sufficiently small, one can write the evolution of $\bm q_t$ as:
\begin{equation}\label{eq:speed_integration_model}
    \bm q_{t+1} = \bm q_{t} + \bm {\dot q}_{t} T_s + (\bm {\dot q}_{t+1} - \bm {\dot q}_{t})\frac{T_s}{2}
\end{equation}
Then, each of the $d_x/2 $ GPs learns the evolution of velocity components by outputting its change $\Delta_t$, and positions are obtained by integration.

\subsection{MC-PILCO}
MC-PILCO is a MBRL algorithm that relies on GPR to learn the system dynamics. The algorithm considers a parametrized policy $\pi_{\bm \vartheta}$ and optimizes the policy parameters $\bm \vartheta$ by minimizing the expectation of the cumulative cost, computed w.r.t. the probability distribution induced by the GP dynamics model. As with most of the MBRL algorithms, MC-PILCO iterates repetitively three main steps: (1) model learning, (2) policy optimization, and (3) policy execution on the system. The model learning phase implements the strategy described in Section \ref{sec:gpr}. In this work, we use the \textit{speed-integration model}. At each execution of the model learning phase, the algorithm updates the one-step-ahead posterior model with \eqref{eq:delta-posterior-mean} and \eqref{eq:delta-posterior-var} using all the interaction data available. The policy optimization phase modifies the policy parameters  $\bm \vartheta$ to minimize the expectation of the cumulative cost, expressed as
\begin{equation}
    J(\bm \vartheta) = \sum_{t=0}^{T} \mathbb{E}_{p_{\bm x_t}}\left[c\left(\bm x_t, \bm u_t\right)\right]. \label{eq:expected-cumulative-cost}
\end{equation}
The expectation is computed w.r.t. the multi-step-ahead state distribution $p_{\bm x_t}$ induced by the one-step-ahead GP dynamics \eqref{eq:one-step-ahead-posterior-mean} and the policy parameters $\bm \vartheta$; the initial state distribution $p_{\bm x_0}$ is fixed and known. Regardless of  $p_{x_0}$, long-term distributions $p_{x_t}$ can not be computed in closed form. MC-PILCO approximates \eqref{eq:expected-cumulative-cost} via a Monte Carlo sampling approach and leverages the reparametrization trick to compute the $\bm \vartheta$ gradient by backpropagation for policy update. At each step of the gradient-based optimization, $M$ particles are sampled from $p_{\bm x_0}$ and simulated for $T$ steps, by selecting inputs according to the policy $\pi_{\bm\vartheta}(\cdot)$, and sampling the next state from the one-step-ahead GP dynamics. The expectations in \eqref{eq:expected-cumulative-cost} are approximated by the particles' cost sample mean. Finally, the gradient obtained by backpropagation is used to update  $\bm \vartheta$. We refer to \cite{MCPILCO} for a detailed description. 

Conveniently, this approximation strategy does not imply any constraints on the GP model and policy definition. The policy adopted in this work is a squashed RBF network expressed as:

\begin{subequations}    
\begin{align}
    \pi_{\bm\vartheta} (\bm x) &= u_\text{max} \tanh{\left( \frac{1}{u_\text{max}} \sum_{i=1}^{n_b} w_i \exp(-\left\| \bm a_i-\bm x \right\|^2_{\Sigma_\pi}) \right)} \label{eq:mcpilco_policy} \\
    &= u_\text{max} \tanh{\left( \phi(\bm x)^\top \bm w\right)}. \label{eq:mcpilco_policy_vect}
\end{align}
\end{subequations}
where \(\bm\vartheta = \{\bm w, A, \Sigma_\pi\}\) are the policy parameters and $u_\text{max}$ is the maximum input applicable (assuming symmetric input bounds).  
The drawback of the Monte Carlo approximation is the considerable computational load. Each simulation costs $O(T \cdot M \cdot N^2)$, where $N$ is the number of training samples. For this reason, limiting the optimization steps is fundamental to reducing actual convergence time.


\section{EXPLORATION-BOOSTED MC-PILCO}
The EB-MC-PILCO framework we propose aims at reducing MC-PILCO convergence time through a smart initialization of the GP dynamics and the policy. To do so, we propose to perform multiple runs of iLQR before starting MC-PILCO. In each run,  we (1) compute control policies through iLQR, (2) apply the control sequence to the system, and (3) train the GPs model on data collected during the rollouts. Then, before starting MC-PILCO, trajectories generated from the final iLQR solution are used to pretrain the MC-PILCO policy parameters within a GPR framework. In the following, we detail (i) the integration of iLQR with GPs, (ii) an exploration strategy we designed for iLQR, and (iii) the policy pretraining procedure. 


\subsection{Integration of iLQR with Gaussian Processes}
To integrate the GP dynamics with iLQR procedure explained in \ref{sec:iLQR} we have to derive \( f_{\bm{x}} \) and \( f_{\bm{u}} \) from GP models. Generally, \( f_{\bm{x}} \) and \( f_{\bm{u}} \) can be computed analytically or estimated using finite differences. For data-driven GP models, finite differences can be computationally inefficient and unstable due to the stochastic nature of GP predictions. A more robust alternative is to compute GP derivatives in closed form. The approach to derivative computation depends on the model type: in a full-state model, each state component dynamics is modeled by a GP, allowing for direct differentiation; in a speed-integration model, GPs predict velocity changes, requiring integration to obtain state derivatives. This problem has been explored in \cite{agpilqr}.
Since iLQR requires a deterministic model, we consider only the GPs posterior means $\mathbb{E}\left[ \Delta_t^{(i)} \right]$, with $i=1,\dots, d_x$. Their analytical derivatives w.r.t the state and input are:
\begin{subequations}
\begin{align}
    \frac{\partial \mathbb{E}[\Delta_{t}^{(i)}]}{\partial \bm{x}_t} = J_x^{(i)} \bm{\alpha}^{(i)} \in \mathbb{R}^{d_{x} \times 1}\label{eq:gp_derivative_state}, \\
    \frac{\partial \mathbb{E}[\Delta_{t}^{(i)}]}{\partial \bm{u}_t} = J_u^{(i)} \bm{\alpha}^{(i)} \in \mathbb{R}^{d_{u} \times 1}\label{eq:gp_derivative_input},
\end{align}
\end{subequations}
where $\bm \alpha^{(i)} = \Gamma_i^{-1}\bm{y}^{(i)} \in \mathbb{R}^{N\times 1}$ is calculated a single time offline during the GP model training. For the SE kernel seen in \eqref{eq:SE_kernel}, the $j$-th column of the state Jacobian $J^{(i)}_{\bm{x}}  \in \mathbb{R}^{d_{{x}} \times N}$, and the input Jacobian $J^{(i)}_{\bm{u}} \in \mathbb{R}^{d_{u} \times N}$ are:
\begin{subequations}
\begin{align}
    J^{(i)}_{\bm x[j]} = & \dfrac{\partial k_\text{SE} ( \tilde {\bm x}_t , \tilde {\bm x}_j) }{\partial \bm{x}_t} =-2 \, k_\text{SE} ( \tilde {\bm x}_t , \tilde {\bm x}_i) \, \Lambda_{\bm{x}}^{-1} (\bm{x}_t - \bm{x}_i),\\
    J^{(i)}_{\bm u[j]} = & \dfrac{\partial k_\text{SE} (  \tilde {\bm x}_t , \tilde {\bm x}_i)}{\partial \bm{u}_t} =-2 \, k_\text{SE} ( \tilde {\bm x}_t , \tilde {\bm x}_i) \, \Lambda_{\bm{u}}^{-1} (\bm{u}_t - \bm{u}_i),
\end{align}
\end{subequations}
where $\tilde{\bm{x}}_j$ with $j \dots N$ is a training input. Then, the linearization of \eqref{eq:transition} is 
\begin{subequations}
\begin{align}
    f_{\bm{x}} = &
    \begin{bmatrix}
    J_{\bm x}^{(1)} \bm{\alpha}^{(1)} & \dots & J_{\bm{x}}^{(d_x)} \bm{\alpha}^{(d_x)} 
    \end{bmatrix}^\top + I_{d_x} ,
   \\
    f_{\bm{u}} = & \begin{bmatrix}
    J_{\bm{u}}^{(1)} \bm{\alpha}^{(1)} & \dots & J_{\bm{u}}^{(d_x)} \bm{\alpha}^{(d_x)}
    \end{bmatrix}^\top,
\end{align}
\end{subequations}
the identity matrix is added to take into account that the GPs model the delta between consecutive states.

\subsubsection*{Speed-integration model}
When using the speed-integration model, the relationship between positions and velocities in \eqref{eq:speed_integration_model} must be leveraged. In this case, in relation to the velocity terms $\dot{\bm{q}}_t$ we can write:
\begin{subequations}
\begin{align}
    A_{\bm{x}} = & \begin{bmatrix}
    J_{\bm x}^{(1)} \bm{\alpha}^{(1)} &  \dots & J_{\bm{x}}^{(d_x/2)} \bm{\alpha}^{(d_x/2)}
    \end{bmatrix}^\top , \\
    A_{\bm{u}} = & \begin{bmatrix}
     J_{\bm u}^{(1)} \bm{\alpha}^{(1)} & \cdots &J_{\bm{u}}^{(d_x/2)} \bm{\alpha}^{(d_x/2)}
    \end{bmatrix}^\top.
\end{align}
\end{subequations}
Then the linearization of \eqref{eq:transition} is expressed by:

\begin{subequations}
\begin{align}
    f_{\bm{x}} &= \begin{bmatrix}
    \frac{T_s}{2}A_{\bm{x}} \\ \\ A_{\bm{x}}
    \end{bmatrix} +
    \begin{bmatrix}
    \bm{0} & T_s I_{d_x/2} \\ \\
    \bm{0} & \bm{0}
    \end{bmatrix} + I_{d_x}, \label{eq:fx} \\
    f_{\bm{u}} &= \begin{bmatrix} \frac{T_s}{2} A_{\bm{u}} \\ \\ A_{\bm{u}} \end{bmatrix}. \label{eq:fu}
\end{align}
\end{subequations}

\subsubsection*{Input Constraints}
Input constraints are not inherently handled by iLQR. As suggested by \cite{ilqr}, several methods can be employed to manage the input constraints. Consistently with MC-PILCO, we use the following squashing function:
\begin{equation}
    \tilde{\bm{u}} = \bm s(\bm u) = \bm u_\text{max} \tanh \left( \dfrac{\bm u}{ \bm u_\text{max}} \right).
\end{equation}
The system dynamics matrix ${f}_{\bm{u}}$ must account for the derivative of $\bm s(\bm u)$ w.r.t $\bm u$. Without loss of generality, let $d_u = 1$. The partial derivative is given by:
\begin{equation}
    \frac{\partial s(u)}{\partial u} = 1 - \tanh^2 \left( \frac{u}{u_{\text{max}}} \right) = \dfrac{u^2_\text{max} - s^2(u)}{u^2_\text{max}}.
\end{equation}
Thus, $f_u$ is updated using the chain rule as $f_u \gets s'(u) f_u$. The squashing function also alters the cost derivatives with respect to the input. If $c(u) = \frac{1}{2} \tilde{R} s(u)^2$, where $\tilde{R}$ is the squashed input cost matrix, then the following hold:
\begin{subequations}
    \begin{align}
        c_u = & \tilde{R} \tilde{u} \frac{\partial s(u)}{\partial u}, \\
        c_{uu} = & \tilde{R} \left( \frac{\partial s(u)}{\partial u} \right)^2 + \tilde{R} \tilde{u} \frac{\partial^2 s(u)}{\partial u^2},
    \end{align}
\end{subequations}
with
\begin{equation}
    \dfrac{\partial^2 s(u)}{\partial u^2} = \dfrac{-2 s(u)}{u^2_\text{max}}\dfrac{\partial s(u)}{\partial u}.
\end{equation}

\subsubsection*{Exploration with iLQR}
iLQR's reliance on a data-driven dynamics model risks divergence due to inaccuracies in the model, especially during initial learning phases. We address this through two complementary strategies:
\begin{itemize}
    \item \textbf{Feedback Compensation}: During the system rollout, we correct deviations from the planned trajectory using:
    \begin{equation}
      \bm{u}_t = \bm{u}_t^* + K_t (\bm{x}_t - \bm{x}_t^*).
    \end{equation}
    where $K_t$ are the feedback gains from the last iLQR backward pass, $\bm{x}_t^*$ is the reference state, and $\bm{u}^*_t$ is the nominal control input.
    \item \textbf{Variance-Adaptive Noise: } We inject Gaussian noise into controls using uncertainty estimates from the GP model, which is fed with the compensated action. The noise variance $\sigma^2$ follows:
    \begin{equation}
        \sigma^2 = \sum_{i=1}^{d_x} \sigma_i^2 + \exp\left(-a\left(\sum_{i=1}^{d_x} \sigma_i^2  - b\right)\right),
    \end{equation}
    where $\sum_{i=1}^{d_x} \sigma_i^2$ is the total prediction variance, $a>0$ controls noise decay and $b$ sets a variance offset. As model confidence increases (lower $\sum\sigma_i^2$), we decrease $a$ to reduce exploration, similar to entropy regularization in reinforcement learning. Implementation details appear in Algorithm \ref{alg:fwd_gaussian_rollout}.
\end{itemize}
The first iLQR iteration is initialized with a random trajectory for exploration. Empirically, we find that initializing subsequent iterations from a zero-control trajectory (instead of warm-starting from prior solutions) avoids overfitting to early suboptimal solutions and improves convergence.

\begin{algorithm}[h]
\caption{Forward Gaussian Rollout Policy}\label{alg:fwd_gaussian_rollout}
\begin{algorithmic}[1]
\renewcommand{\algorithmicrequire}{\textbf{Input:}}
\renewcommand{\algorithmicensure}{\textbf{Output:}}
\REQUIRE $\hat {\bm{x}}, X, \tilde U, K, U, \text{decay rate}, \text{offset}$, step $i \in [0, T/T_s]$
    \STATE Get $i$-th state, actions and gains: $\bm x, \tilde{{u}}, u, K \gets X_i, \tilde{U}_i,  U_i, K_i$
    \STATE Compute unsquashed new input $ g (\hat {\bm x}) \gets u +  K (\hat {\bm x} - \bm x)$
    \STATE Compute squashed new input $\hat{s} =  s( g(\hat{\bm{x}}))$
    \STATE Simulate system with $\hat{\bm x}$ and $\hat{ s}$, retrieve variances $\{\sigma_i^2\}_{i=1}^{d_x}$
    \STATE Compute sum of variances $\sigma_\text{tot}^2 \gets \sum_{i=1}^{d_x}\sigma_i^2$ 
    \STATE Apply  $\sigma_\text{new}^2 \gets \sigma_\text{tot}^2 + \exp(-\text{decay rate} (\sigma_\text{tot}^2 - \text{offset}))$
    \STATE Sample squashed input $\hat {\tilde {{u}}}$ from $\mathcal N(\hat{{s}}, \sigma_\text{new}^2)$
    \STATE Clip squashed input $\hat{\tilde{{u}} } \gets \text{clip}(\hat {\tilde {{u}}}, -{u}_\text{max} + \varepsilon, {u}_\text{max} - \varepsilon)$
    \STATE Recover unsquashed input $\hat { u} \gets {s}^{-1} (\hat {\tilde {{u}}})$
    \RETURN new squashed and unsquashed inputs $\hat {\tilde { u}}$, $\hat { u}$
\end{algorithmic}
\end{algorithm}

\subsection{Pretraining}
To accelerate convergence, we initialize policy parameters using a guiding dataset of trajectories generated by applying iLQR (including feedback/feedforward gains) to the GP model. The dataset inputs are the system's states $X_{iLQR} = \{\bm x_1 ,\dots, \bm x_N\}$, while the outputs $Y_{iLQR}=\{u_1, \dots , u_N\}$ are the corresponding inputs computed by iLQR, namely, referring to the notation introduced in Algorithm \ref{alg:fwd_gaussian_rollout}, $u_i = s(g(\bm x_i))$. When squashing functions are applied, the iLQR inputs must be unsquashed according to the MC-PILCO policy. For instance, in the case of the policy \eqref{eq:mcpilco_policy} :
\begin{equation}
    u_t=\tanh^{-1}\left(\frac{s( g(\bm x_t))}{u_\text{max}}\right).
\end{equation}
Several strategies can be adopted to train the policy. Assuming that the policy is the one in \eqref{eq:mcpilco_policy}, we propose a pretraining based on a Bayesian interpretation. First of all, notice that, extending the compact notation in \eqref{eq:mcpilco_policy_vect} we obtain the following regression problem that can be solved in close form as a linear least squares problem,
\begin{equation}
    Y_{iLQR} = 
    \begin{bmatrix}
        \phi(\bm {x}_1)&
        \dots&
        \phi(\bm{x}_N)
    \end{bmatrix}^\top
    \bm w
    =
    \phi(X_{iLQR})\bm w.
    \label{eq:policy-pretrain-reg}
\end{equation}
We assign to the policy weight $\bm w$ a Gaussian distribution with prior mean $\bm \mu_w$ and covariance $\Sigma_w$, that is $\bm w \sim \mathcal N(\bm \mu_w, \Sigma_w)$. Under this assumption, $\phi(X)\bm w \sim \mathcal N(\phi(X)\bm \mu_w, \phi(X)\Sigma_w\phi(X)^\top)$. Then, in our framework, we pretrain the policy parameters by optimizing the Marginal Log Likelihood instead of directly solving the regression problem \eqref{eq:policy-pretrain-reg}. Conveniently, with this strategy, we can optimize all the policy parameters, including $A$ and $\Sigma_\pi$, and not only $\bm w$. Let $e_\pi := Y_{iLQR} - \phi(X_{iLQR})\bm \mu_w$, then the objective function minimized is 
\begin{align}
    \frac{1}{2}e_\pi^\top \Gamma_{\pi}^{-1} e_\pi + \log|\Gamma_{\pi}|,
\end{align}
where $\Gamma_{\pi} = \phi(X_{iLQR})\Sigma_w\phi(X_{iLQR})^\top + \sigma_{\pi}^2 I$; $\sigma_{\pi}$ is regularization parameter added for numerical stability. The objective is minimized w.r.t. $\bm \mu_w, \Sigma_w, \sigma_{\pi}$ and the policy parameters $A$ and $\Sigma_\pi$. At the end of the optimization, we set $\bm w = \bm \mu_w$.

\section{EXPERIMENTS}
We tested the algorithm on the cart-pole, an under-actuated nonlinear system often used in RL as a benchmark. It consists of a cart that can move horizontally along a track and a pole attached to the cart by a hinge. The objective is to balance the pole upright by moving the cart left or right. The state of the system is represented as \(
    \bm{x} = \begin{bmatrix}p & \dot p & \vartheta & \dot \vartheta \end{bmatrix}^\top
\),  where \( p \) is the horizontal position of the cart, \( \dot{p} \) is its velocity, \( \vartheta \) is the angle the pole forms with the vertical, and \( \dot{\vartheta} \) is its the angular velocity.
The task is to swing-up the pole and stabilize it around $\bm{x}_d = \begin{bmatrix}0 & 0 & \pi & 0\end{bmatrix}^\top$. The running cost for MC-PILCO is defined as:
\begin{equation}\label{eq:mcpilco_cost}
    c(\bm{x}_t) = 1 - \exp \left(- \dfrac{(|\vartheta_t|- \pi)^2}{\ell_\vartheta^2} - \dfrac{p_t^2}{\ell_p^2}\right)
\end{equation}
where $\ell_\vartheta = 3$ and $\ell_p=1$. For iLQR we used a similar cost: 
\begin{equation}\label{eq:ilqr_cost}
    c(\bm{x}, u, \tilde u) = 1 - e^{-||\bm{x} - \bm{x}_d||_{Q^{-1}}^2}+ \frac{1}{2} (\tilde{u}^\top \tilde{R} \,\tilde{u} + u^\top R\, u),
\end{equation}
but with the addition of input costs for both squashed and unsquashed controls. Here, $Q = Q_f = \text{diag}(2, 100, 6, 100)^2$ and $R = \tilde R = 10^{-3}$. Note that while \eqref{eq:mcpilco_cost} admits two solutions (clockwise/counterclockwise swing-up), \eqref{eq:ilqr_cost} has a single basin of attraction. All the experiments use a sampling time of $T_s = 0.05 \, \text{s}$ and a horizon of $T = 3 \, \text{s}$, comprising five interactions with the system, in addition to an initial random interaction. 
To mimic sensor inaccuracies and model imperfections, we inject Gaussian noise with standard deviation $10^{-2}$ into the rollouts.  Success is achieved if $|p_t| < 0.1 \, \text{m}$ and $170 \, \text{deg} < |\vartheta_t| < 190 \, \text{deg}$ for every timestep of the final second of the trial.

\begin{figure}[h]
    \centering
    \includegraphics[width=\linewidth]{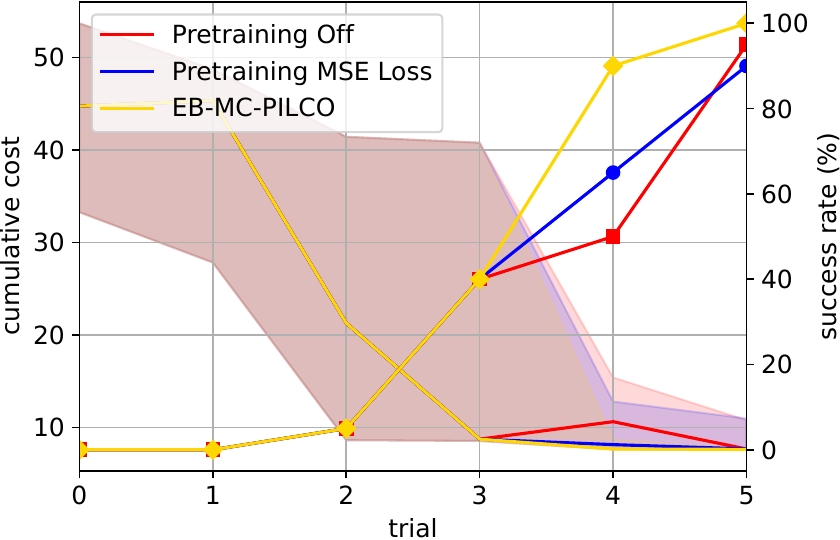}
    \caption{Performance comparison for different pretraining setups under MC-PILCO cost. Solid lines: median values. Bands show 5/95 percentile intervals. Lines with markers: success rate. }
    \label{fig:pretraining_comparison_cost}
\end{figure}

\begin{figure}[h]
    \centering
    \includegraphics[width=\linewidth]{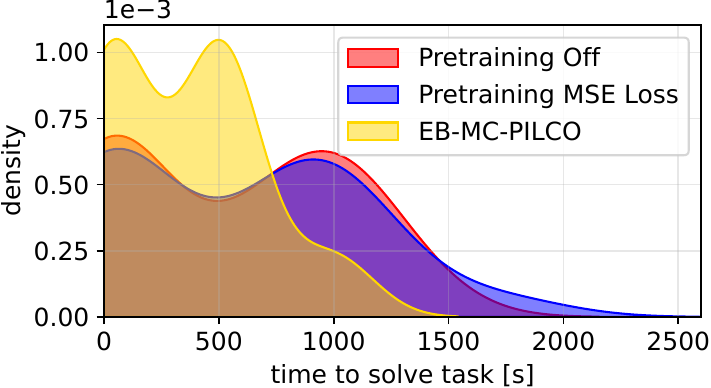}
    \caption{Time comparison for different pretraining setups.}
    \label{fig:pretraining_comparison_time}
\end{figure}

\begin{figure}[h]
    \centering
    \includegraphics[width=\linewidth]{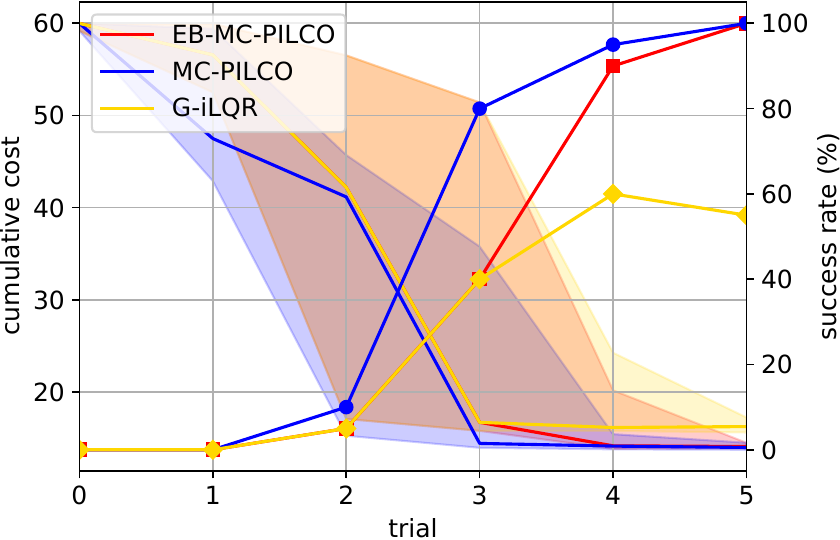}
    \caption{Comparison between EB-MC-PILCO, MC-PILCO and iLQR under PILCO cost. Solid lines: median values. Bands show 5/95 percentile intervals. Lines with markers: success rate.}
    \label{fig:algorithms_comparison_cost}
\end{figure}

\begin{figure}[h]
    \centering
    \includegraphics[width=\linewidth]{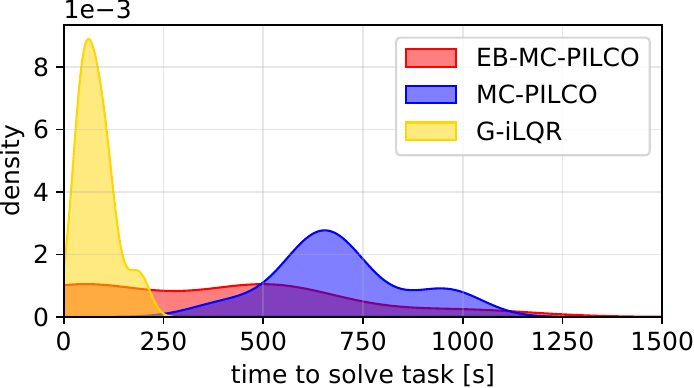}
    \caption{Comparison between methods - Time to solve the task}
    \label{fig:algorithms_comparison_time}
\end{figure}

We fix to three the number of GP-iLQR iterations performed before starting MC-PILCO, as this reliably achieves cost convergence across the median of trials. To assess pretraining efficacy, we compare three configurations that combine iLQR with MC-PILCO using different initialization: (1) no pretraining (iLQR used solely for state-action space exploration), (2) exact MSE loss minimization and (3) the proposed approach. We perform 20 independent runs of each configuration. In each run, the pretraining dataset $D_{iLQR}$ is obtained by simulating 10 trajectories. 
Figure \ref{fig:pretraining_comparison_cost} plots, respectively, the evolution of costs and success rate distributions as a function of trials, while Figure \ref{fig:pretraining_comparison_time} visualizes the distribution of the time to solve the task. EB-MC-PILCO's pretraining achieves two key benefits: (1) it reduces computational overhead while (2) improving the effectiveness  of MC-PILCO's policy optimization phase. Specifically, it decreases the number of optimization steps and yields lower initial costs at Trial 4 (first MC-PILCO iteration) with significantly lower uncertainty.
The bimodal distributions seen in Figure \ref{fig:pretraining_comparison_time} are due to the possibility of solving the task in less than 4 trials, namely using only iLQR.
We performed the same experiment to compare iLQR, MC-PILCO and EB-MC-PILCO. For consistency with previous literature, the cumulative costs are rendered w.r.t. the cost function used in PILCO \cite{deisenroth2011pilco}: 
\begin{equation}\label{eq:pilco_cost}
    c_\text{pilco}(\bm x_t) = 1 - \exp\left(-\dfrac{1}{2} \left(\dfrac{d_t^2}{0.25^2}\right)\right)
\end{equation}
where $d_t^2 = p_t^2 + 2 p_t L \sin(\vartheta_t) + 2L^2 (1+\cos(\vartheta_t))$ is the squared Euclidean distance betweenthe tip of the pole of length $L=0.5\, \text{m}$ and its position at the unstable equilibrium with $p_t = 0 \, \text{m}$. Results are reported in Figure \ref{fig:algorithms_comparison_cost} and \ref{fig:algorithms_comparison_time}. We observe that both MC-PILCO and EB-MC-PILCO achieve 100\% success rate by the last trial, while iLQR  is less robust. 
The comparison between EB-MC-PILCO and MC-PILCO reveals that EB-MC-PILCO solves the tasks significantly faster in most cases, even when it takes more trials ($\geq 4)$. When both solve the task at the fourth trial, EB-MC-PILCO takes $512 \pm 22  \, \text{s}$, while MC-PILCO takes on average $946 \, \text{s}$. Even when MC-PILCO solves the task in 3 iterations, it is still slower than EB-MC-PILCO, with an average of  $ 659 \pm 50 \, \text{s}$.


\section{CONCLUSIONS}
In this work, we have explored techniques to reduce the computational burden of MBRL algorithms, particularly MC-PILCO, by leveraging iLQR for efficient state-action space exploration. While iLQR lacks robustness, it accelerates convergence during the policy optimization phase of MC-PILCO by initializing it.
Results obtained on the cartpole demonstrate substantial improvements in computational efficiency across two scenarios: (1) cases where iLQR alone suffices to solve the task, achieving solutions an order of magnitude faster than MC-PILCO, and, more meaningfully, (2) cases requiring more iterations with the real system, where EB-MC-PILCO still provides significant time savings. These findings highlight the effectiveness of combining faster but less robust methods with MBRL, offering a practical pathway to reduce the computational overhead typically associated with policy optimization in MBRL.
\addtolength{\textheight}{-12cm}   

\bibliographystyle{IEEEtran}
\bibliography{references}
\end{document}